\documentclass{article}

\usepackage{arxiv}

\usepackage[utf8]{inputenc} 
\usepackage[T1]{fontenc}    
\usepackage{hyperref}       
\usepackage{url}            
\usepackage{booktabs}       
\usepackage{amsfonts}       
\usepackage{nicefrac}       
\usepackage{microtype}      
\usepackage{lipsum}
\usepackage{graphicx}
\graphicspath{ {./images/} }

\title{Learning Models for Suicide Prediction from Social Media Posts\thanks{ \ \  This work is accepted to CLPsych 2021,
to be held in conjunction with NAACL 2021}}

\author{
 Ning Wang \\
  Stevens Institute of Technology\\
  Hoboken, NJ 07030 \\
  \texttt{nwang7@stevens.edu} \\
   \And
 Fan Luo \\
  Stevens Institute of Technology\\
  Hoboken, NJ 07030 \\
  \texttt{fluo4@stevens.edu} \\
    \And
  Yuvraj Shivtare \thanks{ \ \   Equal contribution to 2nd author}\\
  Stevens Institute of Technology\\
  Hoboken, NJ 07030 \\
  \texttt{yshivtar@stevens.edu} \\
  \And
    Varsha D. Badal \\
  University of California San Diego\\
  San Diego, CA 92161 \\
  \texttt{vbadal@health.ucsd.edu} \\
  \And
     K.P. Subbalakshmi \\
  Stevens Institute of Technology\\
  Hoboken, NJ 07030 \\
  \texttt{ksubbala@stevens.edu} \\
  \And
    R. Chandramouli \\
  Stevens Institute of Technology\\
  Hoboken, NJ 07030 \\
  \texttt{mouli@stevens.edu} \\
  \And
    Ellen Lee \\
  University of California San Diego\\
  Hoboken, NJ 07030 \\
  \texttt{eel013@health.ucsd.edu} \\

}

\begin{document}
\maketitle
\begin{abstract}
We propose a deep learning architecture and test three other machine learning models to automatically detect individuals that will attempt suicide within (1) 30 days and (2) six months, using their social media post data provided in~\cite{macetal2021} via the CLPsych 2021 shared task. Additionally, we create and extract three sets of handcrafted features for suicide risk detection based on the three-stage theory of suicide and prior work on emotions and the use of pronouns among persons exhibiting suicidal ideations. 
Extensive experimentations show that 
some of the traditional machine learning 
methods outperform 
the baseline with an F1 score of 0.741 and F2 score of 0.833 on subtask 1 (prediction of a suicide attempt 30 days prior).
However, the proposed deep learning method outperforms the baseline with F1 score of 0.737 and F2 score of 0.843 on subtask 2 (prediction of suicide 6 months prior). 
\end{abstract}

\section{Introduction}
\label{sec:intro}
According to World Health Organization (WHO)~\footnote{https://www.who.int/}, close to 800,000 people die due to 
suicide every year, which is one person every 40 seconds. 
The US Centers for Disease Control and Prevention (CDC)~\footnote{https://www.cdc.gov/} claimed that suicide was the tenth leading cause of death overall in the United States.
Recently, there has been a trend in using natural 
language processing (NLP) techniques on 
unstructured physician notes from 
electronic health record (EHR) data to detect 
high-risk patients~\cite{FerEtal18ethical}.

With the proliferation of social media where there is free sharing of information, 
mining data from these platforms has become a natural way to extend the above body of 
work in more natural settings.
Consequently, researchers have started to apply machine learning and NLP 
based techniques to detect suicide ideation on social media platforms \cite{RamEtal20,RoyEtal20}. 
Some of them focused on handcrafted features, including TF-IDF~\cite{WemEtal11}, LIWC~\cite{YlaEtal10}, N-gram, Part-of-Speech (PoS) and emotions~\cite{Faisal201ethical, Ayah191ethical, Lei15ethical, Shaoxiong201ethical}, while others explored language embeddings~\cite{Lei19ethical, Noah19ethical, Ramit18ethical,copper2018glove}.

In this paper, we present several approaches to 
detect suicide ideation from Twitter posts (1) 30 days 
before the attempt and (2) six months before the attempt. We use
the dataset provided by the CLPsych 2021 Shared Tasks~\cite{macetal2021} towards this goal. 
The main contributions of our work are:
\begin{itemize}
\itemsep -2pt
\item Explored and generated multiple handcrafted feature sets motivated by prior work in this area 
\item Proposed a new deep learning architecture that uses latent features from tweets to detect suicide attempts
\item Tested several machine learning algorithms using only handcrafted features and only latent features 
\item Achieved better performance than baseline in terms of F1, F2 and True Positive Rate (TPR) on both subtasks
\end{itemize}

\paragraph{Summary of Findings:} 
The main takeaways from this work are:
\begin{itemize}
\itemsep -2pt
\item Extensive testing on the dataset shows that latent feature (Doc2Vec~\cite{LauEtal16l}), is better at detecting suicide attempts from the tweets than handcrafted features
\item Most of our models performed better on detecting individuals who have attempted suicide or were a victim of suicide than on detecting control individuals who have not
\item The KNN and SVM with latent features perform best on subtask 1, with respect to F1, F2 and TPR; while our proposed C-Attention (C-Att) network performs best on subtask 2, with respect to F1, F2 and TPR
\end{itemize}

\section{Method}
Before we describe the methods in detail we provide a summary of the features used in our work.
We use two classes of features: latent features and handcrafted features. These are described in the sections below.

\subsection{Latent Features}
\label{sec:lat-feat}
Latent features are typically obtained as language embeddings. In our case, we used the Doc2vec~\cite{LauEtal16l} to generate both word embeddings and document embeddings on each post.
Doc2Vec creates a vectorized representation of a group of words
(or a single word, when used in that mode) taken collectively as a single unit. For every document in the corpus, Doc2Vec computes a feature vector. 
There are two models for implementing Doc2vec: Distributed Memory version of Paragraph Vector (PV-DM) and Distributed Bag of Words version of Paragraph Vector (PV-DBOW). For our experimentation, we used Distributed Memory (DM) version. DM randomly samples consecutive words from a sentence and predicts a center word using these randomly sampled set of context words and the feature vector.

\subsection{Handcrafted Features}
\label{sec:HandcraftedFeatures}

\subsubsection{Emotions}
Emotions can be good indicators of depression and suicide ideation~\cite{Bart13ethical,copper2016emotions,Lei20ethical,Soumitra20ethical},
so we include emotions as one of the handcrafted features.
We used the method proposed in~\cite{ShaEtal19} to generate 12 emotion tags, including contentment, pride, fear, 
anxiety, sadness, disgust, relief, shame, anger, interest, agreeableness and joy. Apart from that we also generated 
emotion intensity scores using NRC lexicon~\cite{Moh18}, for the emotions like anger, anticipation, disgust, fear, 
joy, sadness,
surprise and trust. After removing duplicates, we selected 17 emotion tags. 

\subsubsection{Parts of Speech}
We use NLTK~\cite{BirEtal009} to generate Part-of-Speech tags. 
PoS tags can detect the syntactic structure difference between users that  attempt suicide and the control group~\cite{Shaoxiong201ethical}.
It has been shown~\cite{Susan2012ethical} that persons attempting suicide use more first person pronouns.
Therefore, we also calculate the number of occurrences of first person pronouns like ``I", ``me", ``mine" and ``myself" and include this count as another PoS related handcrafted feature.
In total, we generated 34 PoS tags per post for the ``30 days prior prediction" subtask and 37 PoS tags for the ``6 months 
prior prediction" subtask.

\subsubsection{Three-step theory of suicide and suicide dictionary}
We then generate a dictionary of words based on the
three-step theory of suicide (3ST)~\cite{kloEtal15} beginning with the ideation, followed by unmitigated 
strengthening of the idea due to insufficient social support and precipitated by an attempt. These stages are 
underpinned by feelings of hopelessness~\cite{dixetal91}, thwarted belongingness and 
burdensomeness~\cite{chuEtal18,forEtal17}. Violence usually differentiates attempters and 
non-attempters~\cite{staEtal14}. Surviving an attempt is expected to be accompanied by feelings of 
shame~\cite{wikEtal12,wolEtal85}. We expect these feelings to be out of phase with each other creating a leading, 
inline and lagging indicator of suicide attempt. We used Word2vec~\cite{mik1Etal13,mik2Etal13,mik3Etal13} software 
to construct these dictionaries using the accompanying utility (also available in online versions) by evaluating 
closest neighbors of words (gloom and burden, violence, hurt and shame), each containing about 100 words 
with some manual cleanup and editing.
The manual cleanup involved removing stop-words, words with hyphens, special characters, some vernacular tokens, and words that differed in capitalization alone.
We generated this feature set by counting each keyword in each post. In addition, we 
manually created a dictionary of suicide keywords based on suicide-related words published 
in~\cite{LowEtal20,YaoEtal20}, and counted how many suicide-related keywords occurred in each post.~\footnote{Available at:~\url{https://sites.google.com/stevens.edu/infinitylab/suicide-risk-detection}}
\begin{figure}[h] 
\centering
\includegraphics[height=0.56\textwidth]{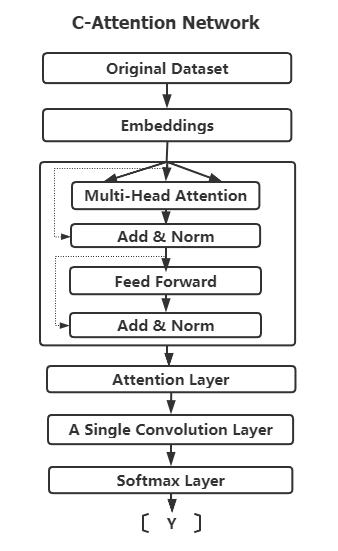}
\caption{The proposed architecture of C-Attention Network}
\label{fig:C_Attention_l}
\end{figure}
\subsection{Models}
In this work, we proposed a deep learning model and used a few other machine learning models for each subtask. 
The proposed deep learning model, which we refer to as the C-Attention Network, is our primary model.
\subsubsection{C-Attention Network}
Figure~\ref{fig:C_Attention_l} depicts our C-Attention network 
which uses latent features to detect suicide attempts. This network is similar to our prior C-Attention Embedding model~\cite{WanCheSub20} with the following differences:
\begin{itemize}
\itemsep -2pt
\item In this work we consider each post as a small document, and use Doc2Vec to generate a 100-dimension embedding representation for each post; whereas the work in \cite{WanCheSub20} generated a sentence 
embedding for each sentence in a speech.
\item We removed the positional encoding layer since there is no positional dependency among posts.
\end{itemize}
In summary, the architecture first calculates the embeddings of the dataset, then processes it via a multi head self-attention (MHA) module that captures the intra-feature relation-ships; an attention layer followed by a single convolution layer and a softmax layer. The MHA module is the same as that proposed in \cite{vasEtal17} for the popular transformer architecture. 

\subsubsection{Latent Features with Other Machine Learning Models}

  In this approach we combined all the posts for each user. Stop words were removed from the posts and lemmatized. 
The average length of posts was found to be 140 words. Long posts were chunked into 150 words segments to retain meaningful information in each post. 
A single $200$-dimension embedding vector is generated for each segment using the Doc2Vec as described in Section~\ref{sec:lat-feat}. 

We applied linear discriminant analysis (LDA)~\cite{mcl2004lda} for
dimensionality reduction before classification. The output of LDA was fed to machine learning models. $K$-Nearest Neighbors (KNN)~\cite{jia2012Knn} with $K$=3, Support Vector Machine (SVM)~\cite{Esteban191ethical} with linear kernel (referred to as SVM(EB) in the rest of the paper) and Decision Tree (D-Tree)~\cite{song2015decision} classifier models were considered.

\subsubsection{Handcrafted Features with Other Machine Learning Models}
We used three other machine learning models on the handcrafted features described in Sec~\ref{sec:HandcraftedFeatures} to address both challenges. The three machine learning models were: Random Forest Classifier (RF)~\cite{Leo01ethical}, Logistic Regression (LR)~\cite{Ahmet18ethical} and Support Vector Machine (SVM)~\cite{Esteban191ethical} (referred to as SVM(HF) for the rest of the paper). 
We used the entire handcrafted features since we found that leaving out any of those handcrafted feature sets would introduce a performance drop.  
We fine-tuned the parameters of each ML model, for example, we set the kernel as rbf (radial basis function) on SVM(HF) model; set the solver as liblinear (limited to one-versus-rest schemes) on LR model; and set the max depth to 4 on RF model to get the best predictions.

\section{Results}
\label{sec:results}
 Table~\ref{tab:results} and Table~\ref{tab:results-new} show the performance results. 
 The results reported in Table~\ref{tab:results} were obtained by running the KNN, SVM(EB) and SVM(HF) models which were 
 trained on the entire training set.
 The performance of the models are measured in terms of 
 F1 and F2 scores, True Positive Rates (TPR), False Positive Rates (FPR) and Area Under the ROC Curve (AUC).
\begin{table}
\centering
\scalebox{0.9}{
\begin{tabular}{lrrrrr}
\toprule
& F1 & F2 & TPR & FPR & AUC \\
\midrule
\multicolumn{6}{l}{\bf Subtask 1 (30 days)}\\
Baseline &0.636 &0.636 &0.636 &0.364 &0.661\\
KNN &0.286 &0.278 &0.273 &0.636 &0.264 \\
SVM(EB) &0.400 &0.377 &0.364 &0.455 &0.529 \\
SVM(HF) &0.364 &0.364 &0.364 &0.636 &0.397 \\
\midrule
\multicolumn{6}{l}{\bf Subtask 2 (6 months)}\\
Baseline &0.710& 0.724 & 0.733 &0.333&0.764\\
KNN &0.429 &0.411 &0.400 &0.467 &0.444 \\
SVM(EB) &0.533 &0.533 &0.533 &0.467 &0.640 \\
SVM(HF) &0.400 &0.400 &0.400 &0.600 &0.502 \\
\bottomrule
\end{tabular}}
\caption{Results obtained by running the KNN, SVM(EB) and SVM(HF) models trained on the entire training set. \label{tbl:results}}
\label{tab:results}
\end{table}
\begin{table}
\centering
\scalebox{0.9}{
\begin{tabular}{lrrrrr}
\toprule
& F1 & F2 & TPR & FPR & AUC \\
\midrule
\multicolumn{6}{l}{\bf Subtask 1 (30 days)}\\
Baseline &0.636 &0.636 &0.636 &0.364 &0.661\\
C-Att & 0.690 &0.806 &\textbf{0.909} &0.727 &0.504 \\
SVM(HF) &0.621&0.726&0.818 &0.818 &0.570 \\
LR &0.571&0.556&0.545 &0.364 &0.434 \\
RF &0.444&0.392&0.364 &\textbf{0.273} &0.603 \\
KNN &\textbf{0.741}&\textbf{0.833}&\textbf{0.909} &0.545 &\textbf{0.694} \\
D-Tree &0.667&0.750&0.818 &0.636 &0.591 \\
SVM(EB) &\textbf{0.741}&\textbf{0.833}&\textbf{0.909} &0.545 &0.653 \\

\midrule
\multicolumn{6}{l}{\bf Subtask 2 (6 months)}\\
Baseline &0.710& 0.724 & 0.733 &0.333&\textbf{0.764}\\
C-Att &\textbf{0.737}&\textbf{0.843}&\textbf{0.933}&0.600&0.76 \\
SVM(HF) &0.600&0.706&0.800 &0.867 &0.518 \\
LR &0.563&0.584&0.600 &0.533 &0.542 \\
RF &0.417&0.362&0.333 &\textbf{0.267} &0.558 \\
KNN &0.500&0.479&0.467 &0.400 &0.536 \\
D-Tree &0.500&0.479&0.467 &0.400 &0.533 \\
SVM(EB) &0.444&0.417&0.400 &0.400 &0.489 \\

\bottomrule
\end{tabular}}
\caption{Results obtained when the training dataset was split into training and validation set as described. HF represents handcrafted features. EB represents word embeddings. \label{tbl:results}}
\label{tab:results-new}
\end{table}
\section{Analysis/Discussion}
\label{sec:analysis}
The results reported in Table~\ref{tab:results} were 
generated by the KNN, SVM(EB) and SVM(HF) models, which 
performed best on the training set. 
From Table~\ref{tab:results}, we can see that
the baseline provided by the CLPsych 2021 shared task
outperformed all of these methods. 

After a
thorough
investigation of the results, we observed that those models that did not perform best on
the training set, performed better on the
test set. It probably indicates that we
over-trained our models on the training set. 

As a result, in the following experiments, we
randomly split the training set into 80\% for training and 20\% for validation, and use the models that performed best on the validation 
set to predict suicide in the test set. The new performance results on the test set are shown in Table~\ref{tab:results-new}.

We noted that in subtask 1, KNN and SVM(EB) performed best in terms of F1, F2 and TRP. 
The best AUC was achieved by KNN only, and the best FPR was achieved by RF.
In subtask 2, C-Att performed best in terms of F1, F2 and TRP; the best FPR was achieved by RF; and the best AUC was achieved by Baseline.\\

Our experiment results would indicate that:
\begin{itemize}
\itemsep -2pt
\item In general, latent features perform better than handcrafted features in this shared task
\item C-Att model performs better on longer range suicide predictions and KNN and SVM(EB) work better on shorter range suicide predictions
\item Besides RF, our other models perform better on detecting suicide individuals than control individuals
\end{itemize}

\section{Conclusion}

In this work, we introduce C-Attention model and 
test other machine learning models to automatically detect suicidal 
individuals based on the latent feature (Doc2Vec) and handcrafted features including emotions, PoS, and three-step theory of suicide and suicide dictionary. Our results show that both KNN and SVM(EB) achieved the best F1 score of 0.741 and F2 score of 0.833 on subtask 1 (prediction of a suicide attempt 30 days prior), and C-Att reached the best F1 score of 0.737 and F2 score of 0.843 on subtask 2 (prediction of suicide 6 months prior).

Ultimately, this work supports the use of social media as an avenue to better predict and understand the experience of suicidal thoughts. However more work is needed to better decipher why certain features and models best predict suicidality in large, diverse, representative samples.

\section*{Ethics Statement}
Secure access to the shared task dataset was provided with IRB approval under University of Maryland, College Park protocol 1642625. 

\section*{Acknowledgements}
\label{sec:ack}
We appreciate the efforts of the organizers
of this challenge to make the data and 
computational resources available to us.

The organizers are particularly grateful to the users who donated data to the OurDataHelps project without whom this work would not be possible, to Qntfy for supporting the OurDataHelps project and making the data available, to NORC for creating and administering the secure infrastructure, and to Amazon for supporting this research with computational resources on AWS.

\bibliographystyle{unsrt}  
\bibliography{alzheimer,references, depression}

\end{document}